\documentclass[letterpaper]{article}
\usepackage{aaai}
\usepackage{times}
\usepackage{helvet}
\usepackage{courier}
\usepackage{dirtytalk}
\usepackage{amsmath}
\usepackage{graphicx}
\graphicspath{ {./images/} }
\begin{document}
\title{Representation Edit Distance as a Measure of Novelty}
\author{Joshua Alspector\\
Institute for Defense Analyses\\
Information Technology and Systems Division\\
Alexandria, VA\\
This research was developed with funding from the Defense Advanced Research Projects Agency (DARPA)\\
Approved for public release; distribution unlimited\\
The views, opinions, and/or findings expressed are those of the author and should not be interpreted\\
as representing the official views or policies of the Department of Defense or the U.S. Government.\\
}
\maketitle
\begin{abstract}
\begin{quote}
Adaptation to novelty is viewed as learning to change and augment existing skills to confront unfamiliar situations. In this paper, we propose that the amount of editing of an effective representation (the Representation Edit Distance or RED) used in a set of skill programs in an agent's mental model is a measure of difficulty for adaptation to novelty.  The RED is an intuitive approximation to the change in information content in bit strings measured by comparing pre-novelty and post-novelty skill programs. We also present some notional examples of how to use RED for predicting difficulty. 
\end{quote}
\end{abstract}

\section{Framework for adaptation to novelty}
\setcounter{section}{1}

\setcounter{subsection}{0}

\subsection{Introduction}

We propose a framework for agents to learn and adapt to novel experiences based on Algorithmic Information Theory (AIT) \cite{Grunwald2007AIT} and the Minimal Description Length (MDL) principle \cite{Rissanen1978MDL}. We chose AIT because it is based on a small set of sound principles, because it is widely applicable to all machine learning methods from traditional AI to deep neural nets, and because instantiations within the framework are falsifiable by testing. Its use depends only on information content and generation, their representation structure, and the probability distributions involved. Agents are viewed as information processing machines (computers) whose experiences add to their information content and skills, and, therefore, their complexity. Complexity in the form of algorithmic entropy is useful for determining difficulty of adaptation to novel experiences. However, because AIT must be applied to optimal and non-computable quantities, one must search for methods that can approximate optimality but are still useful to practically predict adaptation. For this, we hypothesize that a Representation Edit Distance (RED) using near-optimal representations can predict approximate difficulty of adaptation. 

In our view, novelty is in the eye of the beholder. To an agent, any sufficiently advanced technology is indistinguishable from magic \cite{Clarke1973magic}. But if an agent is familiar with the technology, the novelty, and therefore the magic, disappears. In competitions, conflicts, and confrontations, a competitor strives to achieve surprise through novelty to the adversary. The chessmaster prepares a surprising move, the hacker creates a zero-day exploit, and the warrior uses a newly developed secret weapon. The agent's adaptation to novelty is the subject of this paper. Contributions include 1) a framework for using any machine learning method that can adapt to novel experiences; 2) a representation change measure (RED) for determining difficulty of adaptation; 3) a view of novelty as a mismatch between an agent's mental model expectations and its observations; 4) a view of detection, adaptation, and characterization of novelty using ideas in the framework; and 5) some notional examples of how to approximate the framework's representations for prediction testing.

\subsection{A general setting for adaptation}

We consider a primary agent operating in an environment in which novelty is introduced. Novelty can be seen as a reconfiguration of previously existing environmental elements in a new arrangement, or as one or more elements not previously encountered, or as a previously encountered arrangement of elements in a novel context. It may be a situation newly encountered by the primary agent or a novel situation introduced by a secondary agent. The primary agent possesses a mental model of the environment that may include other agents and their mental models (Fig. 1). The mental model includes a model of the environment composed of built-in knowledge and skills (priors in Bayesian terms) and the ability to acquire new knowledge and skills from experience and adapt to novel environmental elements and situations that the agent encounters. Agents have sensors through which they perceive the world, perhaps incompletely and with distortions, and actuators for interacting with their environment. The agent may observe the world passively, enabling the ability, for example, to categorize previously unseen visual object classes. The agent may also be able to act in the world and affect the environment to better adapt to novelty, for example, by moving to get a better view. The ability to adapt to novelty quickly and appropriately is considered a measure of intelligence by some \cite{chollet2019measure}.

\begin{figure}
    \centering
    \includegraphics[width=\columnwidth]{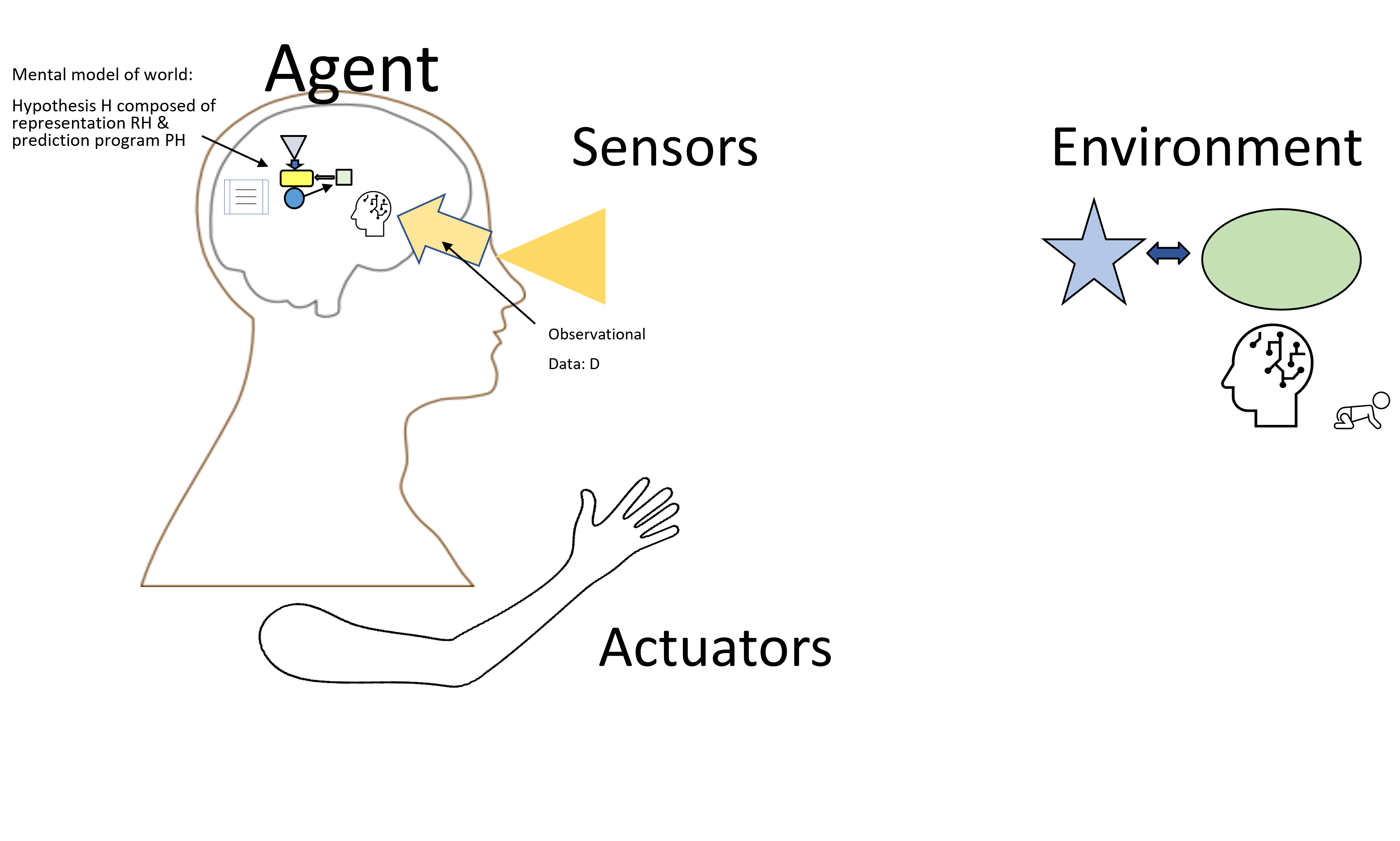}
    \caption{An agent, the environment, and its mental model}
    \label{fig:agent}
   
\end{figure}

The agent senses the elements of the environment through its perceptual system, which is composed of sensors and computational components. The system does not necessarily render the environment perfectly. This imperfect view is modified further by the agent's ability to incorporate the view into its mental model. The mental model view is its interpretation of the environment, which includes an interpretive program for prediction encompassing the agent's skills in the perceived environmental situation. As in Fig. 1, the representation may include distorted or missing entities, attributes, or relations.

We propose a mental model for an agent consisting of a representation portion and a prediction portion. Following the Minimum Description Length \cite{Rissanen1978MDL} or MDL principle, the total description length of these two portions, the \say{structure} of the model, plus a third component describing the observation data, the \say{random noise} of the data deviating from the \say{structure} of the model, should be near a minimum. The representation portion may be, for example, a graph of concepts and relationships and their probability distributions. The complexity of a chosen representation is unlimited as more concepts and relations are learned. An example is the representation of the novel coronavirus, which has become more complex as the world progressed from understanding it as an infectious disease to developing vaccines that require understanding at an atomic level including the complexities of protein folding and human physiology. There are deeper levels of understanding to come. From an Algorithmic Information Theory \cite{Grunwald2007AIT} or AIT perspective, we view a mental model of a situation as a hypothesis, $H$, in a two part form. Part 1 is the representation of the hypothesis, $RH$, such as a function, a neural network, or a probabilistic graphical model. Part 2 is a prediction program that analyzes observational data, assesses the situation, and generates predictions, $PH$. $L(H)$ is the length in bits of the hypothesis description and is the sum of the descriptions lengths of the representation and the prediction program. This is the structural part of MDL. 

Observation data, $D$, is encoded with the help of the hypothesis, $H$, so that $L(D|H)$ is the length, in bits, of the data description for use with the hypothesis model. This is sometimes considered as a noise term. An example would be a model for the prediction of the location of a planet and directions for pointing a telescope at it. An instrumented telescope would record the planet's observation data as a density of light in a location. The density might have a gaussian distribution. If the hypothesis is poor, the light density would not be centered at the predicted location. This would indicate that the hypothesis needs structural change. If the light density is centered at the predicted location but the gaussian spread is larger than the expected parameter encoded in $L(D|H)$, this would indicate that the telescope needs better focus or an improved lens design because it generates a lot of noise in the observation. 

The prediction portion of the mental model can be seen as a program that reads the hypothesis representation and operates on it to generate predictions and, perhaps, select a course of action. The complexity of such a program is approximately fixed in size. One may enforce constant program size by using the same generation/prediction program for all representations of a certain type. This is often the case for knowledge engineering methods. Such a fixed size is applicable to neural nets as well. For example, a neural net inference (prediction) program has the same instructions for all neurons and connections in the network graph. $L(PH)$ is the length in bits of the generation/prediction program, while $L(RH)$ is the length in bits of the representation. $L(H)$ is the sum of the two. According to the \say{crude} two-part MDL principle \cite{Grunwald2007AIT}, a good model minimizes the sum $L(H) + L(D|H)$. Our formulation of an MDL-inspired mental model has three parts because of the two structural portions of the hypothesis (prediction program and representation) plus the noise term describing the data. We argue that the prediction program portion can be ignored in the minimization of code lengths because its code length is close to constant in comparison to the code length of the representation. Thus, once a machine learning and prediction method is chosen, a good representation would be one where its description length plus the description length of observational data is minimized.

\section{Novelty detection, characterization, and adaptation}

\subsection{Detecting novelty}

The prediction program described above operates on a representation that is learned or built-in. This representation may be adapted if the environmental data perceived by the agent does not match the expectation generated by the prediction program. The representation characterizes the perceived state of the world, and if a novel element is introduced or perceived, the mental model adapts its characterization in order to correct its representation and, potentially, improve performance on tasks using the corrected representation. This ability to adapt the representation can be called \textit{characterizing novelty}.  

Novelty and surprise are factors that arouse interest and motivate exploratory or avoidance behavior \cite{Barto2013noveltysurprise}. Novelty has not been previously experienced or encountered, while surprise refers to something unexpected. In our view, the mechanism for novelty and surprise is the same, stemming from an observation that does not match an expectation. This mismatch is a novelty or surprise detection mechanism. \say{Surprise} requires introspection to determine if this entity or situation had been previously encountered. If it has, the mismatch could result from an unexpected context for the situation leading to surprise rather than novelty. Finding a live fish in a desert is an example of surprise, or contextual novelty, but not inherent novelty to the agent.

Prediction errors from novelty and surprise also appear in “predictive coding” architectures \cite{Rao1999predcod}. These are layered systems going from a hierarchy of sensory input levels to levels encoding information in more abstract form. The bottom-up flow of sensory information to abstract representations is paralleled by top-down information predictions to the lower levels. This allows higher-level stages to receive information only through the information mismatch between their predictions and sensations, so that higher levels receive only unexpected information. Our view of novelty encountered by an agent depends on a type of predictive coding within the mental model where the representation can be considered a description code in the MDL sense.

\subsection{Characterizing novelty in representation}

A variety of representations are depicted in Figure 2. In knowledge engineering, a knowledge graph of entities and relations is read by the prediction program and used for inference. In a neural network, nodes and connections create latent representations during the process of learning. These may be difficult to disentangle in the manner of a symbolic knowledge graph, but the simple procedural prediction program produces output inferences just as the program for a knowledge graph does. The third depiction is for regression to a set of data points using, for example, the family of polynomials. Any of these representations plus others such as decision trees, belief nets, and probabilistic graphical models can be cast into a prediction program portion and a representation portion to be minimized along with data for a mental model using MDL techniques. MDL principles \cite{Rissanen1978MDL} can be considered the practical version of AIT \cite{Grunwald2007AIT}. We will motivate our methods using these ideas.

\begin{figure}
    \centering
    \includegraphics[width=\columnwidth]{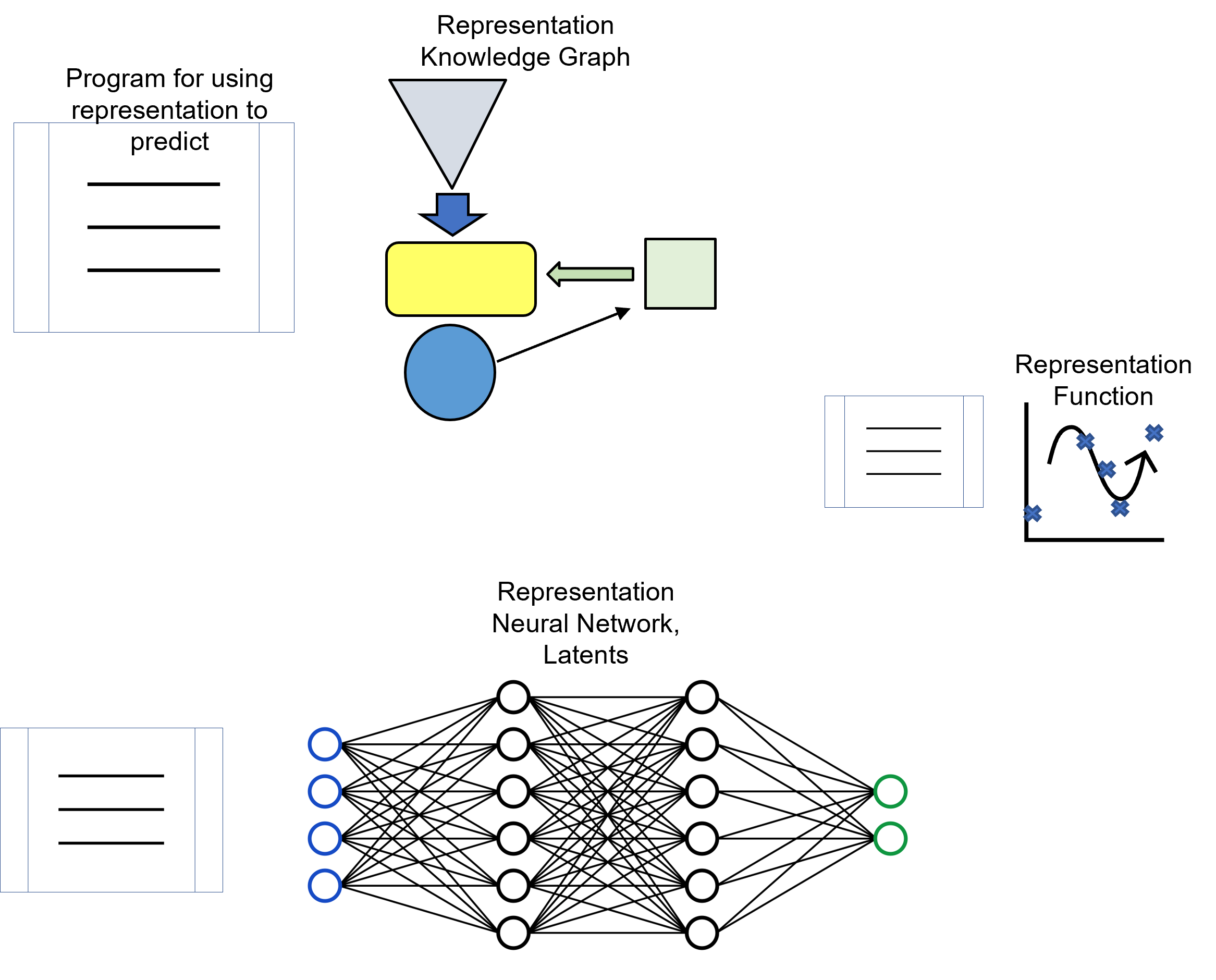}
    \caption{Various machine learning representations and their prediction programs}
    \label{fig:representations}
    
\end{figure}

To adapt its mental model to novelty, the agent can utilize priors, $P$, like prior knowledge and prior mental models along with any new experiences, $E$, it encounters that seem relevant or valuable to such adaptation. Prior models and knowledge add an inductive bias to the adaptation due to the structure of such knowledge and models. For example, if a prior model is in the form of a fully connected neural net, it will not adapt as well as a convolutional neural net (CNN) to a novelty in the visual domain. If the model has an arithmetic and logic unit, as digital computers do, it can be designed to adapt better to knowledge graph inference or first order logic than a neural net will. The difficulty of adaptation depends strongly on the prior model and the task. Adaptation will also depend on the algorithm that optimizes the model based on the representation, priors, and experience. We will call the ability to generalize prior models to a novel situation the \textit{generalization difficulty}, $GD$ \cite{chollet2019measure}, but applied to novelty adaptation rather than a theory of intelligence. 

Regardless of the method used, the change in representation resulting from adaptation can be considered characterizing novelty. If the representation is a knowledge graph, an agent's mental model that adapts to a new entity discovered by observation must add the entity and any relationships observed to its knowledge graph for effective adaptation. Since the beginning of the pandemic caused by the novel coronavirus, the knowledge graph has been augmented by its genetic code, its transmission modes, its shape and the shape of variants, the types of infection observed, the effectiveness of various types of vaccines, etc. If the representation is a neural network, characterization may consist of latent features embedded in the network layers and the connection strengths. For regression, characterization consists of functional forms and parameters. The semantics are clearer in a knowledge graph where natural language is most likely used, but each representation type has its own language and semantics.

\subsection{Approximate information distance}

A useful concept is the information distance between two information objects, say $x$ and $y$, such as computer programs. It is the size of shortest description that transforms each object into the other \cite{Vitanyi2009NIDNCD}. Mathematically, this algorithmic information distance (AID) is the maximum of the conditional Kolmogorov complexities of the transformation programs up to a logarithmic additive term. It can be recast to remove conditional terms.

\begin{equation}
\begin{split}
AID(x,y) = max{[K(x|y),K(y|x)]} \\
= K(xy) -  min[K(x),K(y)]
\end{split}
\end{equation}

It is a universal distance function of shorter length than any other distance function and also a metric. 
Here, $xy$ is the concatenation of the strings. Note that strings $x$ and $y$ may have mutual information to be compressed; for example, they may be a pre-novelty program to be edited for a post-novelty program. To adjust for the AID distance being larger for longer strings, normalize by the maximum of the string lengths to obtain the normalized information distance

\begin{equation}
NID(x,y) = \frac{AID(x,y)}{max[K(x),K(y)]} ,
\end{equation}
which is a metric.

One can use compression algorithms like gzip to measure distances between objects for a practical substitute for the uncomputable Kolmogorov complexity, $K$, of a string using a compressed but approximate version, $Z$. The normalized compression distance is

\begin{equation}
NCD(x, y) = \frac{Z\left(xy\right) -  min\left(Z(x),Z(y)\right)} {max\left(Z(x),Z(y)\right)} ,
\end{equation}
which may be useful for bit-string representations.

One can use the statistics of web search engines to measure distances between semantic ideas, not bit strings. Such concepts come from human knowledge and require semantic context for their meaning. Using the probability that the search term $x$ appears in a page-indexed search engine G, one can derive the normalized web distance, $G(x,y) = NWD(x,y)$, between terms $x$ and $y$ as

\begin{equation}
NWD(x, y) = \frac{[G\left(xy\right)] - min[\left(G(x),G(y)\right)]}{max[\left(G(x),G(y)\right)] } .
\end{equation}

We view the search engine, G, as a compressor using the Web, and $G(x)$ as the binary length of the compressed version of the set of all pages containing the search term $x$. This is not a metric because it does not satisfy the triangle inequality. If concept $x$ (\say{fried}) is semantically close to concept $y$ (\say{chicken}) and concept $y$ is semantically close to concept $z$ (\say{feather}), it is possible that concept $x$ can be semantically very different from concept $z$. The $NWD$ is useful where semantic concepts, such as those in knowledge graphs are the information objects of interest.

\subsection{Information theoretic measures}

\subsubsection{Difficulty of generalizing a skill program to novelty}

The previously mentioned generalization difficulty, $GD$, is a measure of how much the pre-novelty computational mental model must change to adapt acceptably to novelty given a set of experiences in a curriculum, $C$, for learning. It is notionally an edit distance. The concept of adapting acceptably must be defined for each task, $T$, perhaps by a threshold, $\theta$, for a performance measure (e.g., 90 percent correct classification) that the post-novelty computational mental model must reach. Proficiency to reach the acceptable threshold for each task has a value, $T$, representing the task’s subjective importance to the agent expressed as a function, $\omega_T(\theta_T)$.

Algorithmic complexity (AC) measures information content and is also known as Kolmogorov complexity, denoted by $K$. The AC of a binary string, $K(s)$, is the length of the shortest program that outputs the string, $s$. $K(s)$ is not a computable function of $s$, so an approximate measure \cite{Cerra2011RAC} will have to suffice for practical use. This allows approximating $K(s)$ with the compressed version obtained with any lossless compressor $C$ so that $C(s)$ = $K(s)$ + constant. The use of zip-style compression, $Z$, (for strings) and Google-frequency-style compression, $G$, (for ideas) as distance measures is motivated by the need for practical measures.

The relative algorithmic complexity ($RAC$), $K(s_1|s_2)$, is the length of the shortest program that, taking $s_2$ as input, produces $s_1$. The RAC, like the AC, can also be estimated by compression \cite{Cilibrasi2005clustcompress}. To compare the edit distance required to transform one string to another, one must relate the edit distance to the size of some program. For strings of random bits, which have high Shannon entropy, the program size needed, $K$, and the string edit distance are of the same order. If there are regularities in the bit string, such as repeating bit sequences, the program size may be much shorter in a compression code.

Programs can be compared by $RAC$ to determine, for example, how much a pre-novelty program must be edited or adapted to become a post-novelty program. For an agent, a skill program represents a snapshot of its task-specific capabilities including the ability to adapt to novelty within the task. Consider \cite{chollet2019measure}
a task $T$;
$Sol^\theta_T$, the shortest of all possible solutions of $T$ of threshold $\theta$ (shortest skill program that achieves at least skill $\theta$ during evaluation, which may be viewed as the agent's post-novelty program); and
$TrainSol^{opt}_{T,C}$, the shortest optimal training-time solution given a curriculum (e.g., the pre-novelty program that has the best adaptability). The (developer-aware) generalization difficulty ($GD$) is defined as the shortest program that, taking as input the initial system (which may contain built-in priors such as convolutions) plus the shortest possible program that performs optimally over the situations in curriculum $C$, produces a skill program that performs at a skill level of at least $\theta$ during evaluation, normalized by the length of that skill program. Thus, if the developer builds in the solution to the system at $t=0$, the $GD$ is low, while, if the developer relies only on training data without built-in prior bias, the $GD$ is high to reach the solution. $GD$ represents the edits you would have to make to the shortest training-time solution (what training should do) to obtain the evaluation-time solution while using the initial agent’s capabilities without penalty. We will use RED as an approximation to $GD$.

\begin{equation}
GD^\theta_{a,T,C} = \frac{K\left(Sol^\theta_T|TrainSol^{opt}_{T,C, a_{t=0}}\right)}{K\left(Sol^\theta_T\right)}
\end{equation}

Note that this quantity is between 0 and 1 by construction. An approximation to $GD$, like RED, may be used as a measure to judge how difficult a given task would be. If the environment that the agent confronts is a simulated one, a novelty generator can test the adaptation ability $A$ of an agent $a$ to produce a near-optimal skill program. Because such a generator would contain all testable novelties, a developer should be able to produce a program that can adapt well to all of them. Relative complexities can be determined based on the edits needed to turn test agents into the near-optimal agent. We will use MDL to propose a practical, approximate, intuitive version of $GD$, RED, which is computable. 

A task with high $GD$ is one where the evaluation-time behavior needs to differ significantly from the simplest possible optimal training-time behavior in order to achieve sufficient skill. A more generalizable skill program can handle a wider variety of situations than the ones it was trained on. A novelty-adaptable agent is one that can detect, characterize, and adapt to novelty and thus is better able to deal with future uncertainty and novelty.  

\subsubsection{Built-in priors}

Priors (labeled $P$) of an agent, $a$, relative to a task, $T$, and a skill threshold, $\theta$, is the fraction of the AC of the shortest solution of $T$ of skill threshold $\theta$ that is explained by the initial system at the start of the training phase. This does not consider the training data for an optimal adaptation program as $GD$ does but only the baseline initial agent at $t=0$. This is the length (normalized by the $AC$ of the solution) of the shortest possible program that, taking as input the initial agent skill, produces the shortest solution of $T$ that performs at a skill level of at least $\theta$ during evaluation.

\begin{equation}
P^\theta_{a,T} = \frac{K\left(Sol^\theta_T\right) - K\left(Sol^\theta_T\right | a_{t=0}) }{K\left(Sol^\theta_T\right)}
\end{equation}

If the initial system solves the task, the complexity of the solution conditioned on the initial system is 0 and the priors are at the maximum value of 1. If there is no built-in knowledge, the conditional term is high and the same as the solution complexity, so the priors have value 0. \say{Priors} can be interpreted as the \say{amount of relevant information} embedded in the initial system. Priors may be considered as contained in the performance of a baseline pre-novelty-adaptation agent that has no training. Components may include a built-in knowledge base or transfer from pre-training on a similar task or other components of core knowledge. 

\subsubsection{Experience relevant to novelty adaptation}

At each step, the agent is exposed to a situation, responds according to its current program, and receives feedback from the environment thus gaining information that may affect the agent's future responses if the agent is adaptive enough to use it. The increase in experience at time step t is 

\begin{equation}
E^\theta_{a,T,t} = K\left(Sol^\theta_T\right | a_t) - K\left(Sol^\theta_T\right | a_{t-1}, data_t) .
\end{equation}

We can define the experience, $E$, accumulated by an intelligent system about a task during a curriculum as

\begin{equation}
E^\theta_{a,T,C} = \frac{1}{K\left(Sol^{\theta}_T\right)}\sum_t E^\theta_{a,T,t} .
\end{equation}

$E$ measures how much easier it is to create the solution program if the agent makes optimal use of the new information at each step but not necessarily the actual use. Experience can be considered the amount of accumulated information that is novel to the agent and relevant to the task. There may be other information that is not relevant or that is repetitive or redundant to what the agent already knows. The AC of an optimal adaptive agent should increase at each step if the information is novel and relevant, but not if the information is irrelevant. Incremental learners like backpropagation-optimized neural nets may require several exposures to the information to achieve the full increase in experience. The potential uncertainty reduction from the data is that for an optimal learner. Learning curves measure the speed of adaptation to a task from data and depend strongly on the optimization procedure. The increase in AC may be a way to measure novel and relevant information gain if we can approximate $K$ with a computable $C$ from an off-the-shelf compressor.

\subsubsection{Adaptability}

We define the adaptation ability, $A$, of agent, $a$, rather than the intelligence \cite{chollet2019measure} of a system, as a measure of its skill-acquisition efficiency over a scope of tasks, with respect to priors, experience, and generalization difficulty. For a single task, $T$, adaptability can be expressed as the difficulty of the task to the agent divided by the total information contained in the agent's priors and experience. 

\begin{equation}
A^{\theta_T}_{a,T} = \frac{GD^{\theta_T}_{a,T,C}}{P^{\theta_T}_{a,T} + E^{\theta_T}_{a,T,C}}
\end{equation}

This adaptability can be viewed as a generalized sample efficiency that measures how well the agent turns experience and priors into new skills. Put another way, it measures how well an agent adapts to novelty where $GD$ measures the degree of novelty or difficulty. For a scope of tasks on which the agent is tested, the adaptability is averaged over the tasks and weighted by the perceived task value.

\begin{equation}
A^{th}_{a, sc} = \underset {T \in sc}{Avg} \left[\omega_T(\theta_T) \sum_{C \in Cm^{\theta_T}_T} \left[ Pb_c \cdot \frac{GD^{\theta_T}_{a,T,C}}{P^{\theta_T}_{a,T} + E^{\theta_T}_{a,T,C}} \right] \right]
\end{equation}

Sufficient skill threshold, $\theta$, is the subjective threshold of skill associated with a task, above which a skill program can be said to ``solve'' the task $T$.  It is a property of a task. The average over tasks has an implied threshold, $th$. The set of all tasks in scope is $sc$. The probability of a single curriculum is $Pb$. The agent's scope is the space of tasks of non-zero value for which the agent is capable of producing a sufficient solution after a training phase.

$P + E$ (priors plus experience) represents the total exposure of the system to information about the problem, including the information it starts with at the beginning of training. This could be approximated by the information in a set of experiences. This should also include any start-up information either in a baseline system or as a knowledge base that may be used as core knowledge.

The sum over a curriculum subspace, weighted by the probability of each curriculum, represents the expected outcome for the system after training. This assumes that agents are exposed to multiple curricula or training sets in a supervised setting for each task. The subjective value placed on achieving sufficient skill at a task $T$ is $\omega_T(\theta_T)$. The mental model’s value function captures the relative importance of skill at each task subjectively. For humanlike intelligence \cite{Kurdi2019evaluations}, it might be high for tasks relevant to humans, but low for tasks that are not relevant. It would be low for \say{nuisance novelty} \cite{Boult2021unify}, which does not affect task performance.

\section{Representation Edit Distance (RED)}

\subsection{RED as a measure of difficulty}

Prediction error can be the stimulus for learning new concepts and relations. Error may be used to define novelty as a new situation that an agent failed to predict accurately. In the real world, any description of the physical environment is necessarily incomplete and correct prediction is the only test of the mental model and its complexity is the only one that is relevant. An exception would be a simulated world where the model of environmental and agent behavior is fully specified.  Note that environment complexity in a mental model may be reduced by the task the agent must perform because only a portion of the environmental description is needed for executing the task.

We propose RED as an approximate measure for use in determining difficulty. Defining this metric for a particular architecture requires some thought and will depend strongly on the structure of the representation. The size of an effective representation can be approximated by a type of minimum description length such that the representation \say{effectively} expresses the behavior of the system it describes in the mental model. For example, if multiple graphical models, such as random forests, can adequately represent the task and skills needed to a threshold accuracy, and if such a representation can be reduced optimally to a few graphs that are accurate enough, the reduced model can be said to effectively represent the system. We define a near-optimal representation as the minimum size model that can be used to approximate an oracle to a threshold accuracy. Our hypothesis is that edit distances on near-optimal representations can be used for practical information-theoretic measures of difficulty and adaptability. 

\subsection{Complexity for knowledge graphs}

A knowledge graph (KG) \cite{Hogan2021KGs} is a structured representation of facts, consisting of entities, relationships, and semantic descriptions. In Figure 2, an abstract knowledge graph is represented as a set of concept nodes containing semantic descriptions linked by relational edges. KGs may be hand coded or learned, and the representation graph may be embedded in a low-dimensional space. One way the KG may be compressed is to represent each entity in the KG as a vector of discrete codes and compose embeddings from these codes \cite{Sachan2020KGembedcompress}. An approximate complexity may be derived from such techniques. 

KGs are meant to create a semantic model of the environment that may be used to perform inference. Inferences that do not match observation indicate that the KG must be changed or edited. This may be due to the introduction of novel concepts or relations. The edit distance from the previous KG to the new one is a measure of the difficulty. These distance calculations should ideally be performed on the most compressed version of the KG, which would be the Kolmogorov complexity of the graph. For practical use, we posit that a compressed version, perhaps using NWD, that incorporates semantic concepts is an acceptable substitute. Such a description of the environment and the task context should be reduced to its most complete, correct, and consistent but simplest form using a variety of techniques \cite{Hogan2021KGs} for use in determining difficulty.

For an agent, a knowledge and skills graph encompasses concepts, actions, attributes, relations, and other elements useful in characterizing phenomena. An agent's mental model may contain a subset of a larger graph depending on the agent's skill and knowledge level. A useful example is the knowledge and skills needed to adapt to the novel coronavirus. When it first appeared, the previously held knowledge that it is an infectious disease was sufficient to promote the standard mitigation measures of quarantine, disinfection of surfaces, washing hands, and not touching your face. As the virus characterization revealed aerosol transmission, masks and social distancing were added as mitigation measures. The level of granularity for characterization in an ontology of hierarchical concepts had limited success due to less-than-perfect following of mitigation procedures, and it became clear that a much finer granularity of knowledge and deeper characterization would be needed to create vaccines. This involved knowledge at a more atomic level. Knowledge of the molecular structure from the virus genome and protein folding physics revealed the structure in sufficient detail for groups around the world to develop vaccines. Even finer granularity of skills was need for vaccine technologies, especially for the mRNA vaccines. The level of characterization granularity reached is the \say{atomic} level for vaccines or the deepest level of hierarchical and relational knowledge sufficient for executing a task; in this case, developing a vaccine.

Using the refined and reduced semantic knowledge graphs before adaptation and after adaptation to novelty is a matter of replacing the GD of Eq. 5 with the normalized difference in description lengths (code-lengths or compression lengths, C) of the two KGs. 

\begin{equation}
RED^\theta_{a,T,C} = \frac{Cpretr\left(TrSol^\theta_T\right | a_{t=0})[]to]Cpost\left(Sol^\theta_T\right) }{Cpost\left(Sol^\theta_T\right)}
\end{equation}
RED, like GD, represents the edits ($[to]$ in the equation, which may be $Cpost - Cpretr$) one would have to make to the shortest training-time solution (what optimal training should do) to obtain the evaluation-time solution while using the initial agent’s capabilities without penalty on compressed, near-optimal versions of the knowledge graphs. This means that $Cpretr$ must operate near-optimally not only on the pre-novelty data as $Cpre$ does but also on the training curriculum data. However, it can use the capabilities of $Cpre$ without being charged for edit distance on the $Cpre$ graph. Such an optimal use of training data may indicate that $Cpretr$ uses a post-training graph structure. We will posit that the edit distance can be approximated by using the structure of the $Cpost$ graph for the $Cpretr$ graph without the newly learned entities and relations whose edits are charged to RED. 

This difference normalized by the post-novelty codelength is the effective RED in our framework if careful attention is paid to the correct refinement and reduction of the KGs. To determine difficulty of adaptation, we also need to estimate the potential increase in complexity of the agent when it experiences the learning curriculum and adapts optimally. This represents the experience in Eq. 7. For semantic concepts, this may simply be the addition of new concepts and relations to the KG. For example, the change in the structure of a new variant of the coronavirus expressed in atomic form such as a shape description may be called the approximate effective experience, $Eeff$. An optimal agent (an entire drug company plus consultants and knowledge bases) would need just a single exposure to this shape to learn how to make a vaccine if its prior knowledge included vaccine technology. Of course, this prior knowledge, Eq. 6, must also be estimated. For KGs, this is simply the compressed description length, $Cpre$, of the entire pre-novelty knowledge graph. The compressed description length of the priors, $Pd$, according to Eq. 6, would then be the length of the post-novelty solution graph, $Cpost$ minus the length, $Cpre$, divided by $Cpost$.

\begin{equation}
Pd^\theta_{a,T} = \frac{Cpost\left(Sol^\theta_T\right) - Cpre\left(Sol^\theta_T\right | a_{t=0}) }{Cpost\left(Sol^\theta_T\right)}
\end{equation}

Using Eq. 9 and the approximate effective values for priors, experience, and edit distance, the approximate difficulty of adaptation, $Aeff$, to novelty for the task is then

\begin{equation}
Aeff^{\theta_T}_{a,T} = \frac{RED^{\theta_T}_{a,T,C}}{Pd^{\theta_T}_{a,T} + Eeff^{\theta_T}_{a,T,C}} .
\end{equation}

\subsection{Complexity in regression}

MDL is based on finding structure in individual data sequences \cite{Grunwald2007AIT}. Fitting of data to probability distributions (models) extracts regularities viewed as representations that express the \say{structural} meaning of the data. Variations around the structure are the \say{random} portion of the data in the two-part MDL formulation. The MDL principle promotes the idea that for a given set of hypotheses $H$ and data set $D$, one should try to find the hypothesis or combination of hypotheses in $H$ that compresses $D$ most. 

In Fig. 2, a notional polynomial function fits a set of data points. Assume the family of polynomial functions form the set of hypotheses. The perfect fit polynomial (of degree one less than the number of data points) would likely overfit the data and lead to poor generalization to new data from the same source. A third degree polynomial, the \say{model} as shown, seems like a good enough fit, leaving some random variation to the fitted data but able to generalize to new data. The particular fitted parameters, say $y = 3x^2 + 7x + 2$, of the chosen third-degree model form a \say{point hypothesis} in the regression fit. Two-part MDL chooses a point hypothesis that minimizes the sum of description lengths of the model, $L(H)$, plus the data encoded with the help of the model, $L(D|H)$. By using the model of third-degree polynomials to fit the data, we extract the 'structure' of the data in a compressed code leaving only the random errors in the data fit to the hypothesis to encode. If the random portion is distributed according to a simply described function such as a normal distribution, the total code-length would be quite short. If, however, the new data fits a sine function better, the noise portion would be large. A large amount of noise indicates the model needs adaptation to the novel data. This is done by choosing a new function or a new family of functions and a new point hypothesis.

The length of the two-part MDL compressed code is $L(H) + L(D|H)$, and this quantity is to be minimized in a model selection and regression procedure. Difficulty of adaptation to novel data can be estimated by using such a procedure. One may try a family of functions such as polynomials to fit a training set of data points. For the optimization procedure, one would select a degree of the polynomial and perform a regression on the data. The sum, $L(H) + L(D|H)$, would be minimized according to this optimization procedure, which may be a random search guided by a gradient. This would produce a pre-novelty model (the function) whose complexity or code-length can be determined for $Cpre\left(Sol^\theta_T\right | a_{t=0})$. After novelty is introduced or discovered in a test set of data, the model selection and regression procedure is repeated to determine $Cpost\left(Sol^\theta_T\right)$. The function family may be different or the degree of the polynomial may be higher post-novelty. In any case, the priors, $Pd^\theta_{a,T}$, can be determined using Eq. 12. The effective experience, $Eeff^{\theta_T}_{a,T,C}$, can be determined by assessing those data points that are novel and relevant in the test set by measuring the effect of each point on the model parameters by estimating the goodness of fit. The $RED^{\theta_T}_{a,T,C}$ is the edit distance from pre-novelty point hypothesis to post-novelty. We will posit, in analogy with what we did for KGs, that the $Cpretr$ near-optimal function for training is that of the $Cpost$ function before parameter optimization for determining RED in Eq. 11. Putting these quantities together in Eq. 13 determines difficulty of adaptation for the agent to the task of regression for the new data starting from the pre-novelty agent's mental model.

\subsection{Complexity in neural nets}

In one view of neural network function \cite{Tishby2017infoblackbox}, the layers of a deep net form a Markov chain of successive internal representations of the input. The information bottleneck (IB) bound characterizes the optimal representations that maximally compress the input for a given mutual information (MI) on the desired output. A trained deep net processes the input through the chain of hidden layers to the predicted output capturing relevant information in the process. Each hidden layer can be viewed as being an encoder on the input and decoder to the output forming a representation characterized by the layer's MI to the input and to the output regardless of the features encoded. The MI measures the number of relevant bits that the input contains about the output label on average. The optimal learning problem can be seen as the construction of an optimal encoder of that relevant information via an efficient representation. The IB tradeoff between the compression of the input and the prediction of the output leads to an approximate optimal representation. Deep learning effectively finds efficient representations in the hidden layers that are approximately optimal, and this plays a large role in accounting for its success including the surprising lack of overfitting in deep learning. However, this claim and the connection between compression and generalization have been disputed \cite{Saxe2018onIB}.

The problem of compressed models in neural nets has been studied in a supervised setting \cite{Blier2018DLDL}, where inputs to a neural model predict outputs after training. The predictive model should generalize well to inputs not in the training set. Using the MDL principle, one could choose, say, a uniform encoding model with K classes so that probability, $p$, of a class is 1/$K$. The codelength gain, given \say{true} distribution $q$, compared to a uniform code is limited by the amount of mutual information between input and output. There are advantages \cite{Blier2018DLDL} in using a prequential (prediction/sequential) code where one predicts the $n + 1$ label after already having seen $n$ input-output pairs. First, one trains a default model on a few samples of data, then one trains on the resulting encodings. Next, one uses this model for the next few data samples, retrains on all new encodings, and so on. For neural nets, a trainable deep net encoding data incrementally starting from uniform encoding works well even though model parameters are never encoded explicitly. For a convolutional network of depth 8, the authors \cite{Blier2018DLDL} achieve a description length with a compression ratio of 50x, with test set accuracy that is better than that of a variational code \cite{Hinton1993mdlwts}. 

To use these ideas practically requires a compressed coding of a neural network such as a uniform encoding of the synaptic weight parameters. Encodings may reduce the precision of the weights from a floating-point value to a fixed-point value, which, in practice, has been studied extensively, usually in the context of hardware implementations \cite{Gupta2015wtprecision} ; \cite{Hubara2018lowprecision}. The network adjacency matrix for the graph and the binary coded connections form the string to be edited by the optimization method, which may be supervised learning by back-propagation of errors. This learning can take place at full precision if required. We are interested in the compressed version for evaluation of the RED.

The procedure for determining difficulty of adaptation is much the same as in the previous cases discussed. Use the compressed version of pre-novelty and post-novelty networks for evaluation of $Cpre$, $Cpretr$, and $Cpost$ complexities. If, for example, weight dropout is used during training, $Cpretr$ would use the $Cpost$ net structure after training. If no structure change occurs, then $Cpre$ and $Cpost$ network structures are the same. Determine the experience in the training curriculum by estimating the increase in complexity for the agent's neural network representation, assuming the agent can make optimal use of the training examples. A way to do this may be to experiment with batch learning of a representative training sample until the prediction performance saturates. The change in the compressed network representation represents the effective (novel and relevant) experience, $Eeff$. The description length of the priors, $Pd$, is given by Eq. 12. The RED is given by Eq. 11. Eq. 13 then gives the difficulty of adaptation for the agent and task. 

\subsection{Intuitive representations may work well}

In many domains, there may be a multiplicity of representations producing similar accuracy results. This multiplicity has been called \cite{breiman2001statistical} the Rashomon effect after the classic movie by Akira Kurosawa where the same facts of a crime are interpreted in different ways by a multiplicity of characters. A recent study of this effect \cite{Semonova2021rashcurv} noted that a model from a simpler class is approximately as accurate as the most accurate model within the hypothesis space. The authors provide a method for determining when a large set of such equivalent models exist and how likely it is that a simple, interpretable model exists within the class. This supports our complexity analysis of model descriptions and helps to justify the notion that a near-optimal representation, perhaps even a semantic representation, can be edited to provide a RED measure for determining difficulty.

\section{Conclusion and future work}

We have described a framework based on AIT that has applicability to all machine learning methods for adapting to novelty. It remains to test these ideas with approximate instantiations that can provide predictions of difficulty and adaptability using this framework. Methods are needed for good, practical approximations to non-computable Kolmogorov measures within the framework. We propose that RED can be used as an approximate measure to predict difficulty. Research to provide theoretical justifications for methods to produce RED and for its use in prediction is needed. Likewise, it is important to provide experimental tests of predictions of the framework. In conclusion, both theoretical and experimental work is needed to establish and verify the framework proposed.

\section{Acknowledgments}
This work was supported by the Science of Artificial Intelligence and Learning for Open-world Novelty (SAIL-ON) program of the U.S. Defense Advanced Research Projects Agency. It has benefited greatly from discussions with participants in the program.

{\small
\bibliographystyle{aaai}
\bibliography{acffatnovelty}
}

\end{document}